# Waymo Public Road Safety Performance Data

Matthew Schwall   Tom Daniel   Trent Victor   Francesca Favarò   Henning Hohnhold

Waymo LLC

## Abstract

Waymo's mission to reduce traffic injuries and fatalities and improve mobility for all has led us to expand deployment of automated vehicles (AVs) on public roads without a human driver behind the wheel. As part of this process, Waymo is committed to providing the public with informative and relevant data regarding the demonstrated safety of Waymo's automated driving system (ADS), which we call the Waymo Driver.

The data presented in this paper represents more than 6.1 million miles of automated driving in the Phoenix, Arizona metropolitan area, including operations with a trained operator behind the steering wheel from calendar year 2019 and 65,000 miles of driverless operation without a human behind the steering wheel from 2019 and the first nine months of 2020. The paper includes every collision and minor contact experienced during these operations as well as every predicted contact identified using Waymo's counterfactual ("what if") simulation of events had the vehicle's trained operator not disengaged automated driving. There were 47 contact events that occurred over this time period, consisting of 18 actual and 29 simulated contact events, none of which would be expected to result in severe or life-threatening injuries.

This paper presents the collision typology and severity for each actual and simulated event, along with diagrams depicting each of the most significant events. Nearly all the events involved one or more road rule violations or other errors by a human driver or road user, including all eight of the most severe events (which we define as involving actual or expected airbag deployment in any involved vehicle). When compared to national collision statistics, the Waymo Driver completely avoided certain collision modes that human-driven vehicles are frequently involved in, including road departure and collisions with fixed objects. While data related to these collision modes is very promising, the presence of collisions that resulted from challenging situations induced by other drivers serves as a reminder of the limits of AV collision avoidance as long as AVs share roadways with human drivers.

Analysis of events from AV operation on public roads is only one of many complementary safety evaluation methods that Waymo uses, and we are sharing it because it is objective and directly relevant to public road operation of AVs. As automated vehicles continue to improve and fleet mileage continues to grow, so will public understanding of their safety impact. The long-term contributions of this paper are not only the events and mileages shared, but the example set by publicly sharing this type of safety information.

## 1. Introduction

Waymo's automated driving system (ADS), the Waymo Driver, has driven over 20 million miles on public roads since testing began in 2009. Most of these historical miles have been driven with a trained vehicle operator seated in the driver's seat who can take over the driving task, but Waymo also conducts driverless operations in which the Waymo Driver controls the vehicle for the entire trip without a human driver being available to assume any part of the driving task. This form of operation for Waymo's ADS began with the world's first such drive on public roads in 2015 and Waymo is currently expanding driverless operation (without human operators) as part of Waymo's transportation services in the Phoenix, Arizona metropolitan region. By early 2020, these services were providing a combined 1,000 to 2,000 rides per week, 5% to 10% of which were driverless—without any human behind the wheel.

Waymo's safety goal is to reduce traffic injuries and fatalities by driving safely and responsibly. Achieving this goal requires not only superior safety performance by the Waymo Driver, but also public acceptance of automated vehicles ("AVs"). The purpose of this paper is to make available relevant data to promote awareness and discussions that ultimately foster greater public confidence in AVs.

Waymo has published *Waymo's Safety Methodologies and Readiness Determinations* [1], which is an overview of safety readiness methodologies showing how Waymo follows rigorous engineering development and test practices, applying industry standards where appropriate, and developing new methods where those currently available are insufficient. Our extensive experience has taught us that no single safety methodology is sufficient for AVs; instead, multiple methodologies working in concert are needed. These safety methodologies are supported at Waymo by three basic types of testing: simulation, closed-course, and real-world (public road) testing. While each of these forms of testing is a necessary part of Waymo's validation process, public road testing yields some of the most direct measures of the AV's performance within a given operational design domain (ODD). Public road testing and our findings from events experienced in the real world are the focus of this paper. This paper is an example of one of the two Simulated Deployments methodologies discussed in *Waymo's Safety Methodologies and Readiness Determinations* [1].

### 1.1 Public Road Testing

In order to perform initial public road testing of AVs in a safe and responsible manner, trained vehicle operators are seated in the driver's seat and can take over the driving task at any time. Unlike drivers of traditional vehicles, these operators do not execute navigation, path planning, or control tasks. Instead, they are highly trained to focus on monitoring the Waymo Driver's operations within the dynamic environment, and to be prepared to assume control of vehicle operation in the presence of a safety-related conflict. While only a tiny fraction of all disengagements[1] are for reasons related to safety, some do occur in safety-relevant contexts where the vehicle operator disengages automated driving and applies control inputs that alter the vehicle's path so as to avoid a potential collision. Such cautious disengagements improve safety by reducing the risk of collision relative to what the AV may have experienced during initial testing had the AV operated without a human operator. While this provides an immediate reduction of risk during this testing phase, it also means that the outcome of an event experienced by an AV being tested with an operator may not reflect the outcome that the same AV would have experienced if operating in driverless mode, without an operator.

---

[1] In this paper, "disengagement" refers to any event in which the AV's automated driving mode is disengaged. This is broader than the definition used in certain California state regulations, where the term more narrowly refers to certain safety-relevant disengagements.

### 1.2 The Role of Counterfactual ("What If") Simulation

Simulating what might have transpired had a disengagement not occurred is an example of counterfactual ("what if") simulation, a method which is increasingly used (see e.g. [2-5]). The outcomes of counterfactual disengagement simulations are used both individually and in aggregate. Individual counterfactual disengagement simulations provide an opportunity to study what would likely have occurred in a specific scenario had the Waymo Driver remained engaged. Waymo analyzes each disengagement to identify potential collisions, near-misses, and other metrics. If the simulation outcome reveals an opportunity to improve the behavior of the ADS, then the simulation is used to develop and test changes to software algorithms. The disengagement event is also added to a library of scenarios, so that future software can be tested against the scenario. At an aggregate level, Waymo uses results from counterfactual disengagement simulations to produce metrics relevant to the AV's on-road performance.

Counterfactual disengagement simulations can be significantly more realistic than simulations that are created entirely synthetically because they use the actual behavior of the AV and other agents up to the point of disengagement. Simulation is used to represent the predicted vehicle response for a brief period (seconds) after disengagement, and the simulation outcome provides insight into what could have happened had the trained operator not intervened. Still, these simulations carry the limitation of being informative, but not definitive. Furthermore, as with crash simulations and human body models, simulations of the behavior of human agents are an area of ongoing refinement. Waymo's models will continue to evolve, and even for these brief simulations, future models may result in different simulated outcomes.

### 1.3 Aims and Contributions of this Paper

The aim of this paper is to share information about driving events that is informative about the safety performance of the Waymo Driver. This paper provides information about collisions and other, more minor, contacts experienced during Waymo's public road operations. This paper includes safety data in the form of event counts and event descriptions from over 6.1 million miles of driving conducted in the Waymo Driver's driverless ODD. This mileage figure represents over 500 years of driving for the average U.S. licensed driver [6]. For these miles, this paper provides information regarding (1) every actual contact event that vehicles were involved in during driverless operation with and without trained operators, as well as (2) events in which the vehicle's trained operator disengaged and subsequent counterfactual simulation resulted in any contact between the AV and the other agent, had the disengagement not occurred.

## 2. Methods

### 2.1 Data sources

Waymo first operated a driverless vehicle without an operator on public roads in 2015 in Austin, Texas, but the base vehicle platform and ODD have changed since this time. The data in this paper is from Waymo's current AVs capable of driverless operation, which are built on the Chrysler Pacifica platform and operate in driverless mode on public roads within an ODD in the East Valley region of the Phoenix, Arizona metropolitan area. The ODD includes roadways with speed limits up to and including 45 miles per hour. Driverless operations occur at all times day and night, except during inclement weather including heavy rain and dust storms.

The data shared in this paper comes from:
- *Driverless* operation, in which the automated driving system controls the vehicle for the entire trip without a human driver behind the wheel or otherwise being available to assume any part of the driving task. The data from driverless mode shared in this paper is from January 1, 2019 to September 30, 2020, during which Waymo AVs drove 65,000 miles in driverless mode in the metro Phoenix ODD.[2]
- *Self-driving with trained operators*, in which the automated driving system controls the vehicle but there is a trained vehicle operator in the driver's seat who can disengage and take over the driving task. The data from self-driving with trained operators mode shared in this paper is from January 1, 2019 to December 31, 2019, during which Waymo AVs drove 6.1 million miles in this mode in Waymo's metro Phoenix ODD.[3]

### 2.2 Data from Actual Collisions and Minor Contacts

The data in this paper includes every actual collision and minor contact event that occurred involving a Waymo AV operating in driverless mode or self-driving with trained operators mode in Waymo's metro Phoenix ODD. For the purpose of this paper, we have chosen to include every event that involved contact between the AV and another object. This definition encompasses not only every severity of collision, but also events such as a pedestrian walking into the side of the stationary AV.

### 2.3 Data from Counterfactual ("What If") Simulation

This section discusses the simulation and analysis processes used to determine which instances of operator disengagements would likely have resulted in contact with other road users had the vehicle continued in its automated operations. This involves simulation first of the AV, then if needed, of the behavior of other agents.

**Simulation of the AV motion post-disengagement**

After a vehicle operator disengages, the manually-controlled trajectory of the Waymo vehicle will likely differ from the one the AV would have followed had the disengagement not occurred. The first step in post-disengage simulation is therefore to simulate the AV's counterfactual post-disengage motion. This is performed by providing a simulation running Waymo self-driving software with the AV's pre-disengage position, attitude, velocity, and acceleration along with the AV's recorded sensor observations and simulating the response of the software and resulting motion of the Waymo vehicle.

These post-disengage simulations can be performed using compatible versions of the AV's software, which is useful for comparing the performance of different software versions with the same scenarios. However, for the results presented in this paper, the simulations of the Waymo Driver's post-disengage behavior were performed using the software version that was running on the Waymo Vehicle at the time of the disengagement. A consequence of this is that the data presented in this paper is primarily representative of the Waymo Driver's performance in 2019, and does not represent the latest performance of the Waymo Driver. Results could have been presented using the 2019 disengagements and the most recent version of the Waymo Driver's software, but to do so could be misleading, since the most recent software benefited from learnings gleaned from the 2019 disengagements.

---

[2] Driverless mode mileage and data begins January 1, 2019 to align with the beginning of the period used for self-driving with trained operators mode data and ends September 30, 2020 to include the only collision that Waymo has experienced in this mode, which occurred in September 2020.

[3] Calendar year 2019 was selected to provide a full year of data, thereby controlling for potential seasonality effects. Data from 2020 has not been included, as much of this operation occurred during the COVID-19 pandemic, during which traffic conditions in Waymo's metro Phoenix ODD were not representative of other time periods.

After the AV's post-disengage motion is simulated, a check is performed to determine if the simulated positions of the AV overlap at any point with the recorded positions of other agents. Overlapping positions indicate a potential collision, but do not necessarily indicate that a collision would have occurred. This is because, in these initial simulations, the other agents are modeled to have the same behavior as was observed following the driver's disengagement. This may not be realistic in cases where the other agents would likely have responded differently to the AV's counterfactual simulated motion than they did to the AV's actual post-disengage motion. In such cases, further simulation is required.

As an example, consider a hypothetical case in which the AV, with a vehicle following behind it, detects a pedestrian on the sidewalk near a crosswalk and the AV begins slowing in response. The vehicle's operator determines that the pedestrian is not attempting to cross the street and disengages and resumes the AV's initial speed. Post-disengage simulation reveals that the AV's software would have slowed significantly for the pedestrian before later proceeding. The vehicle following behind the AV did not slow during the actual event, because the AV operator's rapid disengagement made slowing unnecessary. Therefore, the initial post-disengage simulation might show that the vehicle behind the AV overlaps with the AV's simulated position at some points in time. However, this is potentially unrealistic since had the AV operator not disengaged, the driver in the following vehicle would likely have slowed for the slowing AV. This example illustrates a case where a realistic post-disengage counterfactual simulation requires also modeling the behavior of other agents.

**Modeling of other agents**

Models of other agents are necessary for counterfactual simulations in which the counterfactual behavior of the AV may have elicited a different response from other nearby roadway users. Fortunately, while modeling agent behavior over long periods of time is challenging, understanding plausible conflict-avoidance or collision-avoidance behavior over the short time horizon following a disengagement is a more feasible task. Waymo expresses short-term agent responses using human collision avoidance behavior models. These models aim to capture the responses of human drivers, motorcyclists, cyclists, and pedestrians to collision avoidance situations, such as braking by a lead vehicle or being cut-off by another agent who fails to yield right-of-way. Because only the agent's short-term response needs to be modeled, the space of plausible reactions to such stimuli can be defined using a discrete set of factors such as response times to specific inputs and brake or swerve ability.

Though the plausible space of responses to a driving situation is limited, even factors as (seemingly) simple as response time vary between drivers placed in identical scenarios due to variables such as human performance differences and momentary differences in visual attention to the driving task. Just as these human differences can determine whether a collision on the road occurs or is avoided, when reflected in simulation models, these differences can alter whether a simulation results in a simulated contact or not. Waymo considers a broad spectrum of potential human driving performance in developing and evaluating the AV, but for transparency and simplicity, the results reported in this paper are based on deterministic models that generate a single response to a given input. Other methods can be used to capture a range of possible human responses, such as probabilistic counterfactual outcomes, but they are more complex.

Waymo's proprietary human collsion avoidance behavior models are based on existing road user behavior modelling frameworks [7,8] and calibrated using naturalistic human collision and near-collision data. The agent's collision avoidance actions are modeled as occurring in response to deviations between the agent's initial expectations and how the situation actually played out (i.e., violations of the agent's expectations [7]). The agent's response is further constrained by human braking and steering limitations. Waymo uses different models for different types of agents, including heavy trucks, pedestrians, and cyclists, and for different stimuli such as a forward agent braking or an agent emerging from behind an occlusion.

Human collision avoidance behavior models are employed for disengagements in which there is overlap between the simulated post-disengage trajectory of the AV and the actual post-disengage trajectory of another agent. In these cases, instead of using the agent's recorded post-disengage trajectory, the post-disengage trajectory of the other agent is determined by applying the relevant human collision avoidance behavior model. In the prior example of the AV slowing for a pedestrian near a crosswalk, a human collision avoidance behavior model would be used to determine the simulated behavior of the vehicle behind the AV. The AV's motion as it slows defines the stimulus to the following driver's collision avoidance model. The output of the model is a simulated braking and/or swerving response by the following driver after the modeled response time.

**Contact analysis of simulated collisions**

After the human collision avoidance behavior models have generated the simulated trajectories of other agents, these trajectories are compared with the simulated post-disengage trajectory of the AV to determine if simulated contact occurs. Our simulation analysis indicates that disengagements would rarely result in contact. In fact, in more than 99.9% of disengagements, no simulated contact is found to occur. In the rare cases where contact is inferred, the event is analyzed to determine the event severity of the resulting contact. This determination categorizes collisions based on likelihood of injury and is based on the collision object (e.g., other vehicles, static objects, or vulnerable road users such as pedestrians or cyclists), impact velocity, and impact geometry. Waymo's methods for determining event severity category are developed using national crash databases and are periodically refined to reflect updated data.

| Row# | Event type | Manner of Collision ("Other" = non-Waymo vehicle) | S0 | S1 (no airbag deployment) | S1 (airbag deployment any vehicle) | S2 | S3 | Collision % Contribution US * | Fatal Collision % Contribution U.S. (Maricopa Cnty, AZ) ** |
|---|---|---|---|---|---|---|---|---|---|
| | | | Waymo-involved collision-relevant contacts by ISO 26262 severity classification Actual & simulated event counts (Totals in Bold) | | | | | Human Crash Statistics (Non-Waymo Data) | |
| 1 | Single Vehicle Events | Road Departure, Fixed object, Rollover | 0 | 0 | 0 | 0 | 0 | 20% | 27% (21%) |
| 2 | | Striking a pedestrian/cyclist | 0 | 0 | 0 | 0 | 0 | 2% | 33% (41%) |
| 3 | | Struck by pedestrian/cyclist | 1 (actual) 2 (sim) | 0 | 0 | 0 | 0 | <0.5% | 1% (1%) |
| 4 | Multiple Vehicle Events | Reversing | 1 (actual) 1 (sim) | 0 | 0 | 0 | 0 | 1% | <0.1% |
| 5 | | Other reversing, Waymo straight | 1 (actual) 1 (sim) | 0 | 0 | 0 | 0 | | |
| 6 | | Waymo reversing, Other straight | 0 | 0 | 0 | 0 | 0 | | |
| 7 | | Sideswipe (Same Direction) | 1 (actual) 8 (sim) | 1 (sim) | 0 | 0 | 0 | 11% | 1% (1%) |
| 8 | | Other lane change, Waymo straight | 1 (actual) 7 (sim) | 0 | 0 | 0 | 0 | | |
| 9 | | Waymo lane change, Other straight | 1 (sim) | 1 (sim) | 0 | 0 | 0 | | |
| 10 | | Head-on + Opposite Direction Sideswipe | 0 | 0 | 1 (sim) | 0 | 0 | 5% | 9% (7%) |
| 11 | | Rear End | 11 (actual) 1 (sim) | 1 (actual) 1 (sim) | 2 (actual) | 0 | 0 | 34% | 5% (5%) |
| 12 | | Other striking, Waymo struck (stopped) | 8 (actual) | 0 | 0 | 0 | 0 | | |
| 13 | | Other striking, Waymo struck (slower) | 2 (actual) | 0 | 1 (actual) | 0 | 0 | | |
| 14 | | Other striking, Waymo struck (decelerating) | 1 (actual) | 1 (sim) | 1 (actual)† | 0 | 0 | | |
| 15 | | Waymo striking, Other struck (stopped) | 0 | 0 | 0 | 0 | 0 | | |
| 16 | | Waymo striking, Other struck (slower) | 0 | 0 | 0 | 0 | 0 | | |
| 17 | | Waymo striking, Other struck (decelerating) | 1 (sim) | 0 | 0 | 0 | 0 | | |
| 18 | | Angled | 4 (sim) | 6 (sim) | 1 (actual) 4 (sim) | 0 | 0 | 27% | 24% (24%) |
| 19 | | Same direction - Other turns across Waymo straight travel | 0 | 2 (sim) | 0 | 0 | 0 | | |
| 20 | | Same direction - Other turns into Waymo straight travel | 3 (sim) | 0 | 2 (sim) | 0 | 0 | | |
| 21 | | Opposite direction - Other turns across Waymo straight travel | 0 | 0 | 1 (sim) | 0 | 0 | | |
| 22 | | Opposite direction - Other turns into Waymo straight travel | 0 | 0 | 1 (sim) | 0 | 0 | | |
| 23 | | Straight crossing paths | 0 | 1 (sim) | 1 (actual) | 0 | 0 | | |
| 24 | | Same direction - Waymo turns across other straight travel | 1 (sim) | 3 (sim) | 0 | 0 | 0 | | |
| 25 | | Same direction - Waymo turns into other straight travel | 0 | 0 | 0 | 0 | 0 | | |
| 26 | | Opposite direction - Waymo turns across other straight travel | 0 | 0 | 0 | 0 | 0 | | |
| 27 | | Opposite direction - Waymo turns into other straight travel | 0 | 0 | 0 | 0 | 0 | | |
| 28 | | Total | 14 (actual) 16 (sim) | 1 (actual) 8 (sim) | 3 (actual) 5 (sim) | 0 | 0 | 100% | 100% (100%) |

*CRSS 2016-2018, Urban area, ≤45 mph roadways
**FARS 2016-2018, Urban area, ≤45 mph roadways
† denotes sole collision in driverless operation (without human operator present)

Table 1 - Classification of Waymo-involved collisions (6.1 M AV miles driven)
(Rightmost columns are non-Waymo data derived from NHTSA collision databases involving Class 1 vehicles)

## 3. Results: Collisions and Minor Contacts

Table 1 provides an overview of all collision and other, more minor, contact events in the data sources, categorized in rows according to their collision typology using the *Manner of Collision* categories from National Highway Traffic Safety Administration (NHTSA) collision databases such as the Fatality Analysis Reporting System (FARS) [9], and subcategories similar to other NHTSA coding variables. The Waymo-involved events are tallied in columns categorized by estimated event severity using the ISO 26262 [10] severity classes: S0, S1, S2, and S3, ranging from no injury expected (S0) to possible critical injuries expected (S3). This scale is based on likelihood of AIS injury level [11] (e.g. S1 signifies at least 10% probability of AIS-1 level or higher level injury), which Waymo has estimated for both actual and simulated collisions using the change in velocity and principle direction of force estimated for each involved vehicle.

In order to provide more information about event severity within the S1 designation, S1 severity events have been separated into two columns in Table 1 based on whether each event is of sufficient severity to result in actual or simulated airbag deployment for any involved vehicle.[4] Of the eight airbag-deployment-level S1 events, five are simulated events with expected airbag deployment, two were actual events involving deployment of only another vehicle's frontal airbags, and one actual event involved deployment of another vehicle's frontal airbags and the Waymo vehicle's side airbags.[5] There were no actual or predicted S2 or S3 events.

The rightmost two columns of Table 1 include human collision statistics in the form of the percent contributions from each *Manner of Collision* to the total counts of collisions and fatal collisions. These values have been calculated using NHTSA's Crash Report Sampling System (CRSS) (police-reported collisions) and FARS [9] (fatal collisions only) data for collisions in urban land zones occurring on ≤45 mph roadways, which approximates Waymo's ODD. They serve as an informative reference to understand the breakdown of contributions by collision typology for non-Waymo data. Comparison between these human collision statistics and Waymo event counts provides insight into the Waymo Driver's opportunity for reducing injuries and fatalities due to collisions, and is discussed further in Section 4.

In total, the Waymo vehicle was involved in 20 events involving contact with another object and experienced 27 disengagements that resulted in contact in post-disengagement simulation, for a total of 47 events (actual and simulated). In two of the actual events (which occurred after disengagement), post-disengage simulation revealed that the event would have been moderately more severe had the trained operator not disengaged. Therefore, these two events are treated in this paper according to their more severe simulated outcomes, yielding a total of 18 actual outcomes and 29 simulated outcomes.

The following sections are ordered to describe the events according to groups with similar characteristics within each *Manner of Collision.* In addition to these group descriptions, diagrams are provided for all of the most severe or potentially severe events. Specifically, diagrams have been provided for every actual or simulated event in which a pedestrian or cyclist[6] was involved (three events) and every event with actual or simulated airbag deployment for any involved vehicle (eight events).

### 3.1 Single Vehicle Events

The *Manner of Collision* categories within the NHTSA crash database can be broadly classified as either single vehicle events, which involve a single motorized vehicle in transport,[7] or multiple-vehicle events, which involve the impact of at least two motorized vehicles in transport.

The single-vehicle *Manner of Collision* category has been parsed here into three subcategories: (1) events that involve road departure, contact with the roadway environment/infrastructure or other fixed objects, or single-vehicle rollover (2) events involving the motorized vehicle striking a pedestrian or bicyclist, and (3) events involving the motorized vehicle being struck by a pedestrian or bicyclist. The first two of these groups of single-vehicle collisions combine to contribute approximately 60% to all human-driven fatal collisions on ≤45 mph urban roadways, both nationally and within Maricopa County, Arizona, where Waymo's ODD is located.

The Waymo Driver did not have any events (actual or simulated) in this data that involved road departure, contact with the roadway environment/infrastructure or other fixed objects, or rollover. There were also no collisions (actual or simulated) in which the Waymo Driver struck a pedestrian or cyclist. There were three events (one actual, two simulated) in which the Waymo vehicle was struck by a pedestrian or cyclist. In each instance, the Waymo Driver decelerated and stopped, and a pedestrian or cyclist made contact with the right side of the stationary Waymo vehicle while the pedestrian or cyclist was traveling at low speeds. These three events are illustrated below and none of these actual or simulated events can reasonably be considered injurious.

To summarize the single-vehicle event outcomes, the Waymo Driver was involved in one actual and two simulated non-injurious events where a pedestrian or cyclist struck a stationary Waymo vehicle at low speeds.

---

[4] The severity threshold used to differentiate more severe from less severe S1 events is actual airbag deployment in any involved vehicle or, for simulated events, expected airbag deployment (e.g. for front airbags, a longitudinal crash pulse deltaV ≥ 10 mph) associated with any involved vehicle. Eight of the seventeen events of S1 severity events meet this threshold and the remaining nine are listed as *S1 (no airbag deployment)* in Table 1.

[5] Care should be taken if comparing these events with other data sources (e.g. naturalistic driving data), as other data sources may report airbag deployment only in a subject vehicle, rather than any involved vehicle.

[6] There were no events involving motorcyclists or other powered two-wheelers.

[7] This category is captured in the NHTSA coding manual as "Not a Collision with a Motor Vehicle in Transport."

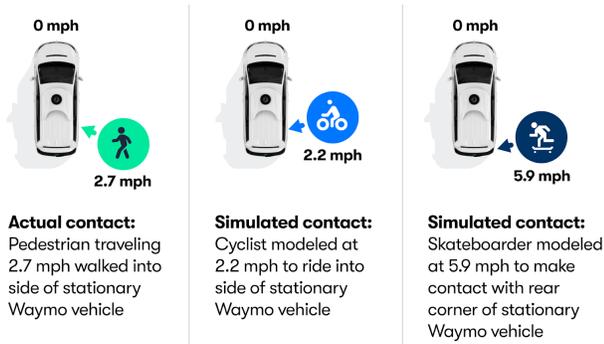

Figure 1. All actual and simulated contact events involving pedestrians or cyclists

## 3.2 Multiple Vehicle Events: Reversing

Reversing collisions[8] (e.g., rear-to-front, rear-to-side, rear-to-rear) are usually associated with parking lot events or occur on local (≤ 25 mph) roadways and do not frequently appear in databases of police-reported crashes. There were two such collisions involving the Waymo Driver, one actual and one simulated (both S0 severity). In both scenarios, the Waymo vehicle was stopped or traveling forward at low speed and the other vehicle was reversing at a speed of less than 3 mph at the moment of contact to the side of the Waymo vehicle.

## 3.3 Multiple Vehicle Events: Same Direction Sideswipe

Same direction sideswipe collisions are a more common vehicle collision mode, and are typically low in severity. These events are typically experienced during lane changing or merging maneuvers. The Waymo Driver was involved in ten simulated same direction sideswipe collisions. The events in this category have been assigned to the subcategories in rows 8 and 9 of Table 1.

**Other vehicle changing lanes, Waymo vehicle straight**

The collisions in this subcategory involved seven simulated collisions and one actual collision. In each of these, the Waymo vehicle was stopped or traveling straight in a designated lane at or below the speed limit. The other vehicle changed lanes into the area occupied by the Waymo vehicle, which resulted in simulated or actual sideswipe collisions.

---

[8] Reversing collisions can include "Rear-to-rear", "Rear-to-side", and "Other" in NHTSA databases' definition of Manner of Collision.

**Other vehicle straight, Waymo vehicle changing lanes**

The collisions in this subcategory involve two simulated collisions. In both of these simulations, the Waymo Driver made a lateral movement in front of a vehicle traveling straight in an adjacent lane. In one of the events, the following vehicle was traveling over 30 mph above the posted speed limit. The other event involved a vehicle that had entered early into a dedicated left turn lane that the Waymo Driver was attempting to merge into.

## 3.4 Multiple Vehicle Events: Head-on or Opposite Direction Sideswipe

Head-on collisions have the potential for high severity outcomes and, as shown in Table 1, contribute approximately 9% of the human-driven crash fatalities nationally on ≤ 45 mph urban roadways.

The data includes one event in this category, which occurred when the Waymo vehicle was traveling straight in a designated lane while self-driving with a trained operator late at night. This event involved another vehicle traveling the wrong direction in the Waymo vehicle's lane of travel (see Figure 2 below, Event A[9]).

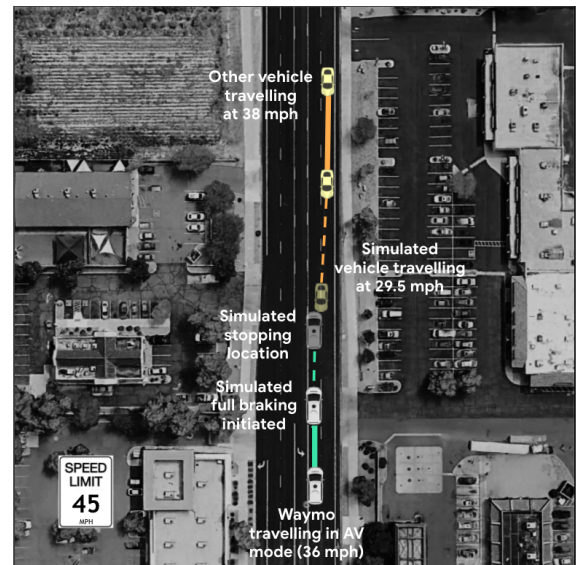

Figure 2. Event A - Simulated head-on collision involving the other vehicle traveling the wrong way

---

[9] Within the diagrams of Figures 2 to 9, actual collisions are represented in color, while simulated ones feature a black and white background. Solid trajectory lines represent those observed in real life, while dashed trajectories and shaded poses represent simulated conditions. Diagrams are intended for visual reference only, and are not drawn to scale.

In simulation, the Waymo Driver detected the wrong way vehicle, initiated full braking, and was simulated to come to a complete stop in its lane prior to impact. The simulated collision assumes that the wrong way vehicle would have continued on the same path as observed in the actual event. The absence of simulated collision avoidance movement by the other vehicle reflects our assumption based on driving behavior and circumstances that the other driver was significantly impaired or fatigued. The resulting simulated collision shows the other vehicle traveling 29 mph when it strikes the stationary Waymo vehicle (S1 severity with expected airbag deployment).

### 3.5 Multiple Vehicle Events: Rear End

Rear End collisions are the most common collision type in human-driven collisions (Table 1), though they contribute to only approximately 5% of the human-driven collision fatalities in the US and in the Waymo ODD. As shown in Table 1, the Waymo Driver was involved in fourteen actual and two simulated rear end collisions. In all but one of these events, another vehicle struck the rear of the Waymo vehicle. Rows 12 to 17 of Table 1 list these events in subcategories based on NHTSA variables indicating whether the lead vehicle was stopped, moving slower, or decelerating at the time of collision. For additional context regarding these events, the text below describes the events using four groupings that cut across these subcategories and are based on the actions leading up to event onset.

**Rear end struck event group, Waymo vehicle stopped or gradually decelerating for traffic controls or traffic ahead while traveling straight**

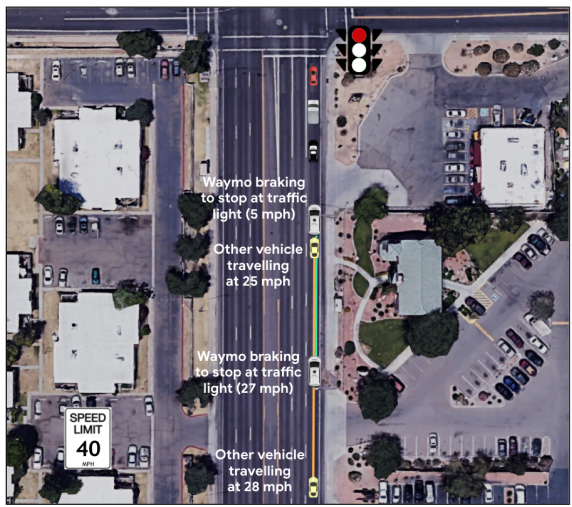

Figure 3. Event B: A rear end collision that resulted in airbag deployment for the vehicle that struck the Waymo vehicle. Sole collision in driverless mode, without a trained operator in the driver's seat.

This grouping consists of eight actual collisions (including two of S1 severity) from the subcategories in rows 12-14 of Table 1. In these collisions, the Waymo vehicle had been traveling straight and was stopped for a traffic control device (six cases) or gradually decelerating (two cases) due to traffic controls or traffic conditions. Most (six cases) of these collisions had relative contact speeds less than 6 mph. Figure 3 (Event B) depicts the one collision within this grouping that involved actual or expected airbag deployment (S1 and resulting in airbag deployment of the striking vehicle).

To summarize, in this group of eight actual rear end struck events, the Waymo was struck while stopped or gradually decelerating for traffic controls or traffic ahead.

**Rear end struck event group, Waymo vehicle moving slower while traveling straight**

Two actual collisions involved the Waymo vehicle being struck on the rear bumper while traveling straight at a constant speed at or below the speed limit. In one collision, the Waymo Driver had slowed to a constant speed in the course of traveling over a speed bump. In the other collision (Figure 4, Event C), the Waymo vehicle, traveling straight at the speed limit, was struck by a vehicle traveling 23 mph over the posted speed limit. Both collisions were of S1 severity, with airbag deployment occurring in the striking vehicle in the latter collision within this grouping.

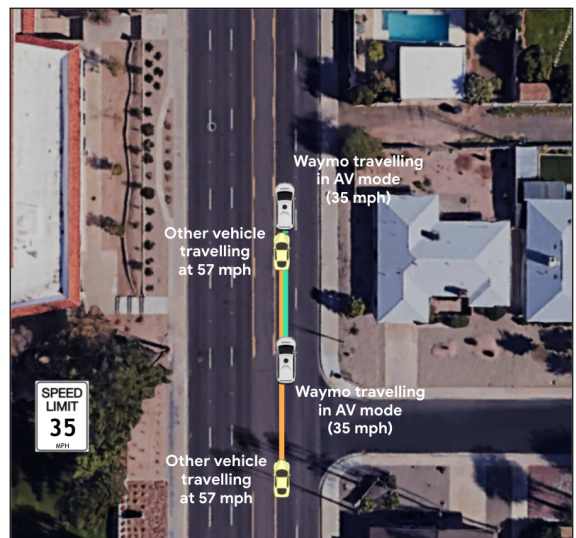

Figure 4. Event C: A rear end collision that resulted in airbag deployment for the vehicle that struck the Waymo vehicle.

**Rear end struck event group in right turning maneuvers**

Four actual crashes involved the Waymo vehicle being struck on the rear bumper at right turns or slip lanes. These collisions occurred while the Waymo was stationary or near stationary waiting for crossing traffic to clear after having gradually slowed to account for this traffic. Relative impact speeds for all 4 events were less than 6 mph.

**Rear end struck event with braking of lead vehicle during left turn**

The remaining rear end struck collision involved a deceleration to a near stop by the Waymo Driver while making a left turn in an intersection with a following vehicle that was traveling at a speed and following distance that did not allow for the following driver to successfully respond to the Waymo Driver's braking. The simulated collision impact was estimated to be 16 mph, and this event is categorized as S1 severity.

**Rear end striking event**

The single simulated event (row 17 in Table 1) in this grouping involved a vehicle that swerved into the lane in front of the Waymo and braked hard immediately after cutting in despite lack of any obstruction ahead (consistent with antagonistic motive). The Waymo Driver was simulated to have achieved full braking in response to the other vehicle's braking, but was simulated to contact the lead vehicle with a relative impact speed of 1 mph (S0 severity).

## 3.6 Multiple Vehicle Events: Angled

Angled collisions, those that are typically seen at intersections and involve crossing or turning vehicles, account for approximately one quarter of all human-driven collisions and a similar fraction of the contribution to all human-driven fatalities. The Waymo Driver was involved in fourteen simulated and one actual angled collision. Rows 19 to 27 of Table 1 list these events in subcategories based on NHTSA variables indicating movement of the vehicles (same direction, opposite direction, or straight crossing paths) and the relative geometry of the turning motion. These events also can be described using the two groupings below.

**Angled event group with the other vehicle not yielding to Waymo right-of-way**

This grouping consists of the events in rows 19 to 23 of Table 1. The collisions in this grouping (ten simulated, one actual) involve the Waymo vehicle traveling straight in a designated lane at or below the speed limit. In all scenarios, the turning/crossing other vehicle either disregarded traffic controls or otherwise did not properly yield right-of-way.[10]

---

[10] Right-of-way is determined based on the positions of vehicles prior to contact with respect to the intersection geometry, roadway markings, and the status of traffic control devices. Right-of-way is useful as a means of categorizing some events, but it can be insufficient to determine collision responsibilities since it does not reflect all road rule violations (e.g. speeding), nor does it provide information regarding collision avoidability. In order to avoid collisions, the Waymo Driver recognizes that yielding even when the Waymo vehicle is entitled to right-of-way may be more appropriate to decrease the risk of collision, for example when encountering an incautious other agent.

Consistent with the elevated severity of these collisions in human-driven vehicle statistics, five of the eleven collision-relevant contacts in this grouping would likely result (or did, in the one actual collision) in airbag deployment for at least one of the involved vehicles. Diagrams of those events are shown below in Figure 5 - 9.

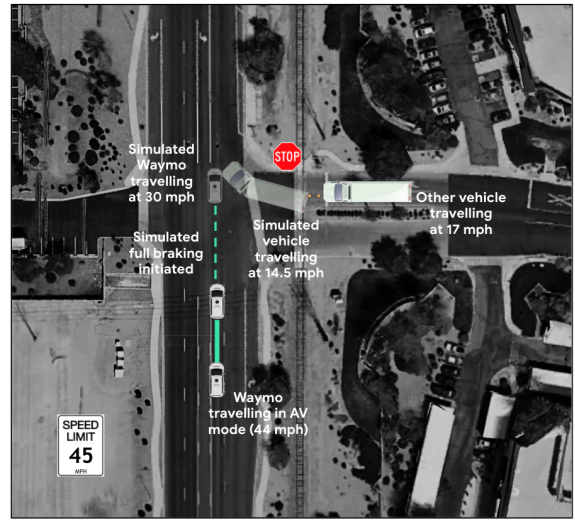

Figure 5. Event D - Simulated Waymo traveling straight with other vehicle failing to stop at a stop sign and not yielding right-of-way to Waymo vehicle

Events D and E (Figures 5 and 6) involve collisions to the side of the Waymo vehicle that were brought upon by another vehicle that failed to obey the applicable traffic control device. In Event D (Figure 5), the tractor-trailer failed to stop at the stop sign, and the simulated Waymo Driver, predicting that the tractor-trailer would enter its lane of travel, enacted full braking to avoid collision, reducing its speed from 44 mph to 31 mph. Event E (Figure 6) was an actual collision involving a vehicle that ran through a red light while the Waymo vehicle was proceeding through the intersection crossing with a green light.

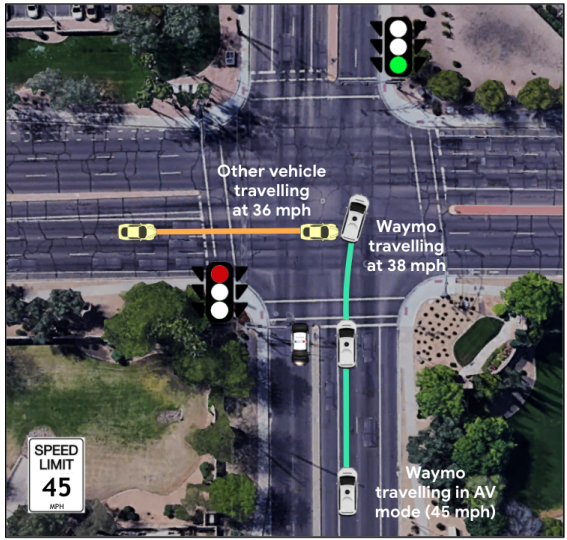

Figure 6. Event E - A collision in which the other vehicle passed through a red stop light

Figures 7 and 8 (Event F and G) depict simulated angled collisions with vehicles that failed to yield right-of-way when entering the Waymo vehicle's lane of travel. In both events, the line-of-sight between the Waymo and the other vehicle was occluded prior to the simulated collision. In both instances, when the simulated Waymo Driver became aware of the other vehicle's intention to enter the travel lane, the simulated Waymo Driver initiated braking in an attempt to avoid/mitigate impact.

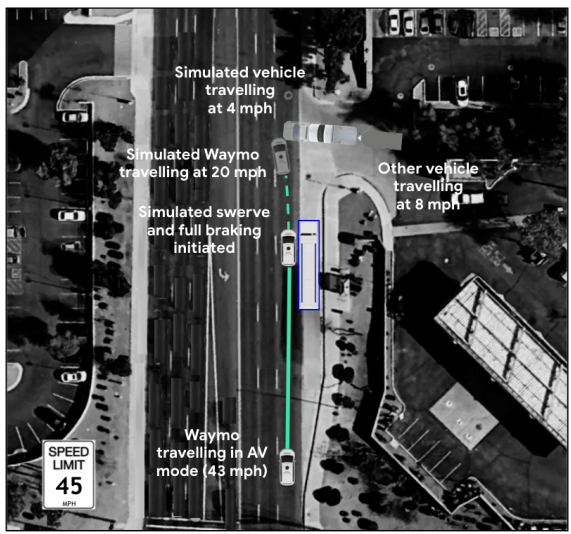

Figure 7. Event F - Simulated collision involving crossing vehicle behind occlusion

The Waymo Driver's simulated response in Event F resulted in a 23 mph simulated speed reduction prior to impact and also involved initiation of an evasive swerve. In Event G, there was insufficient time for a significant reduction in speed prior to simulated collision.

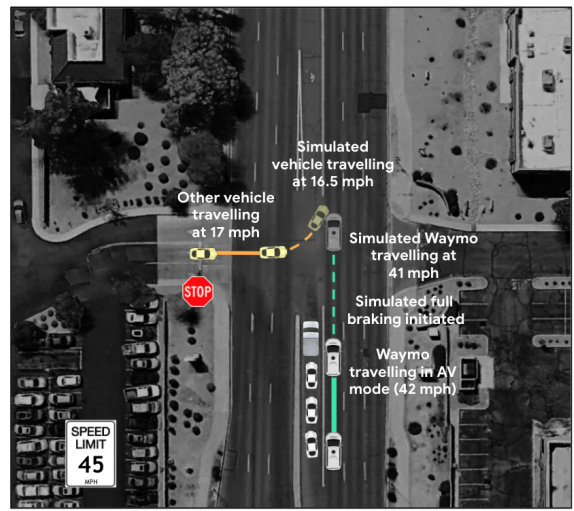

Figure 8. Event G - Simulated collision involving crossing vehicle behind occlusion

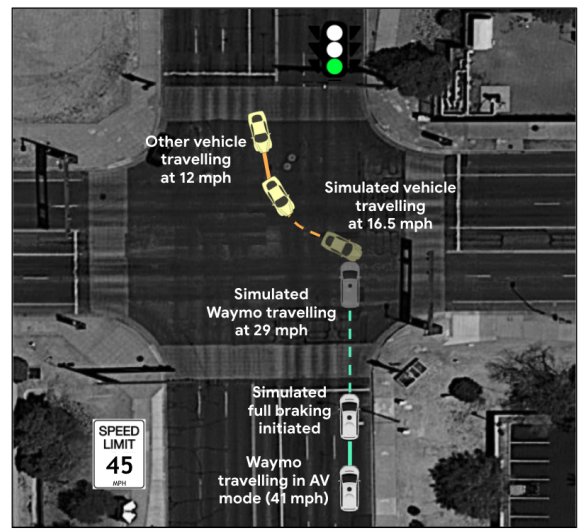

Figure 9. Event H - Simulated collision involving other vehicle not yielding to Waymo right-of-way

The simulated collision in Figure 9 (Event H) depicts a vehicle making a left turn across the Waymo vehicle's travel path. The other vehicle did not have the right-of-way at any point leading up to the depicted sequence of events. The Waymo Driver's simulated response to the vehicle's action was the initiation of braking just prior to entering the intersection. Simulated full braking was achieved, resulting in a 12 mph speed reduction before simulated impact. Based on the vehicle masses and simulated vehicle speeds and geometry at impact, this event is categorized as S1 with expected airbag deployment. It is the most severe collision (simulated or actual) in the dataset and approaches the boundary between S1 and S2 classification.

In sum, the collisions in this angled event grouping (ten simulated, one actual) were characterized by the other vehicle failing to yield right-of-way to the Waymo vehicle, and they involved the Waymo vehicle traveling straight with the right-of-way, with a speed at or below the speed limit.

**Angled event group with Waymo vehicle crossing another vehicle's path**

The collisions in this grouping (row 24 of Table 1) involve four simulated collisions, where the Waymo Driver was making a right turn from a rightmost lane that was either splitting to an additional lane, or had been the result of two lanes merging to one. In each event, a passenger vehicle attempted to pass the Waymo vehicle on the right while the Waymo Driver was slowing to make the right turn with the right turn signal activated. In each case, the Waymo vehicle's trained operator disengaged, while in simulation the Waymo Driver turned, resulting in simulated collision. Using NHTSA coding variables, these are categorized as a right turn across the path of a following vehicle.

In sum, this angled event group involved four simulated events where a vehicle attempted to pass the Waymo vehicle on the right as the Waymo Driver was preparing for and making a right turn.

# 4. Discussion

The purpose of this paper is to make available information about driving events that is relevant to the safety of the Waymo AVs currently operating in driverless mode on public roads. The goal of this transparency is to contribute to broad learning with the industry, policymakers, and the public; promote awareness and discussions; and foster greater public confidence in automated vehicles. To summarize the findings from the data above:

- Over 6.1 million miles of automated driving, including 65,000 miles of driverless operation without a human behind the wheel, there were 47 collision or other contact events (18 actual and 29 simulated, one during driverless operation).
- Of the sixteen **rear end** events, eight events involved the Waymo being struck while stopped or gradually decelerating for traffic controls or traffic ahead. Two events involved the Waymo being struck while traveling at a constant speed. Another group of five rear end struck events were characterized by inadequate response by other vehicles to the Waymo vehicle's slowing behavior when turning. The single event where the Waymo was the striking vehicle involved a passing vehicle that swerved into the lane in front of the Waymo vehicle and braked hard despite lack of any obstruction ahead (consistent with antagonistic motive).
- Of the fifteen **angled** events, eleven events were characterized by the other vehicle failing to properly yield right-of-way to the Waymo vehicle traveling straight at or below the speed limit. Another group of four angled events involved vehicles attempting to pass the Waymo vehicle on the right as the Waymo Driver was slowing to make a right turn.
- Of the ten **sideswipe** (same direction) events, eight events involved another vehicle changing lanes into where the Waymo vehicle was stopped or travelling straight in a designated lane at or below the speed limit. Two sideswipe events occured when the Waymo Driver made a lateral movement in front of a vehicle in an adjacent lane (in one case the other vehicle was travelling over 30 mph above the posted speed limit).
- All of the three **single vehicle** events were non-injurious (S0 severity) events which involved the Waymo vehicle being **struck by a pedestrian or cyclist** while stationary.
- Two **reversing** events (S0 severity) involved other vehicles reversing at < 3 mph into the side of a Waymo vehicle, while the Waymo vehicle was either stopped or traveling forward below the speed limit.
- One **head-on** (S1 severity) event occurred with another vehicle traveling the wrong direction at night in the Waymo vehicle's lane of travel, after the Waymo Driver had stopped in reaction to the oncoming vehicle.

## 4.1 Collision Avoidance: Management of Human-Driver-Related Contributing Factors

Humans exhibit a large variation of driving behaviors including deviations from traffic rules and safe driving performance that can lead to collisions. Nearly all events summarized above involved one or more road rule violations or other driving performance deviations by another road user. Table 2 below lists these contributing factors for the eight most severe or potentially severe events depicted in Figures 2-9.[10]

| Event Identifier | Other road user's road rule violation or other contributory action |
|---|---|
| A | Driving on the wrong side of the road |
| B | Failure to yield to a slower moving vehicle traveling in the same direction |
| C | Failure to yield to a slower moving vehicle traveling in the same direction, speeding (57 mph in a 35 mph zone) |
| D | Failure to stop at a stop sign, failure to yield to a vehicle approaching from the left while making a right turn at an unsignalized intersection. |
| E | Passing through a stale red stop light at 36 mph |
| F | Failure to yield to a vehicle approaching from the left while making a left turn from an unsignalized commercial driveway |
| G | Failure to yield to a vehicle approaching from the right while making a left turn from an unsignalized commercial driveway |
| H | Failure to yield to an oncoming vehicle during an unprotected left turn at a signalized intersection |

Table 2 - Actions of the other road user in each of the events involving airbag deployment or likely airbag deployment.

The contribution of human deviations from safe driving practices observed in Tables 1 & 2 underscore the importance of collision-avoidance behaviors in achieving Waymo's goal of reducing injuries and fatalities due to collisions. In addition to Waymo's key focus on not causing collisions, Waymo also works to mitigate possible collisions due to human behaviors such as inattention, aggressive driving, and speeding. The events that resulted in contact and are presented here represent only a tiny fraction of the incautious behaviors encountered in 6.1 million miles of driving. Although many of these situations would not be present in a future with a high proportion of AVs, we envision sharing roads with human drivers for the foreseeable future. The rare contact events described in this paper are used to develop enhanced collision avoidance to improve traffic safety, and we will continue to focus on enhancing avoidance of human-induced collisions.

Beyond collision avoidance, Waymo also continually investigates improvements to the Waymo Driver's behaviors to reduce the likelihood of conflict with human-driven vehicles and other road users. For example, in each of the 3 actual and simulated events in which a pedestrian or cyclist struck the Waymo vehicle at low speed, the Waymo vehicle had decelerated and stopped immediately prior to the contact or simulated contact in a way that may have differed from the cyclist's or pedestrian's expectations. This illustrates a key challenge faced by AVs operating in a predominantly human traffic system and underscores the importance of driving in a way that is interpretable and predictable by other road users.

The primary purpose of Waymo's public road operations is to continue refining and improving AV operations in their intended environment. Unlike human drivers, who primarily improve through individual experience, the learnings from an event experienced by a single AV can be used to permanently improve the safety performance of an entire fleet of AVs. As a result, AV performance can continually improve, while aggregate human driving performance is essentially stagnant.

## 4.2 Aggregate Safety Performance

The mix of events in Section 3 highlights certain performance characteristics of the Waymo Driver. The Waymo Driver experienced zero actual or simulated events in the "road departure, fixed object, rollover" single-vehicle collision typology (Row 1 in Table 1, 27% of all US roadway fatalities), reflecting the system's ability to navigate the ODD in a highly reliable manner. In addition, while rear-end collisions are one of the most common collision modes for human drivers (rows 12 to 17 in Table 1), the Waymo Driver only recorded a single front-to-rear striking collision (simulated) and this event involved an agent cutting in and immediately braking without allowing for adequate separation distance (consistent with antagonistic motive).

**Lower-severity collision risk**

In both human-driven and automated vehicles, lower-severity events (S0 and S1) occur at significantly higher frequency than higher-severity (S2 and S3) events. As a result, fewer miles are needed to draw statistical conclusions about S0 and S1 rates. When comparing driving data, the mileage needed to reveal statistically significant differences also depends on the magnitude of the differences in the actual rates being compared. For a given metric, the larger the difference in performance, the fewer miles that are required to establish statistical confidence in a hypothesis of non-inferiority or superiority. The 6.1 million miles in self-driving with trained operators mode underlying the data in Section 3 provide sufficient statistical signal to detect moderate-to-large differences in S0 and S1 event frequencies, and Waymo makes use of these event rates for tracking longer-term improvements to the Waymo Driver.

**Higher-severity collision risk**

6.1 million miles does not provide statistical power to draw meaningful conclusions about the frequencies of events of severity S2 or S3. At this mileage scale, the statistical noise is extremely large and zero or low event counts only provide performance bounds, which necessitates the consideration of other metrics to fully assess the safety of the Waymo Driver. As a consequence, Waymo uses other methods to evaluate the higher-severity performance, including both simulation-based and closed-course scenario-based collision-avoidance testing [1], In addition, low-severity data, when evaluated in the context of each event's collision geometry, may be informative of high-severity risk. While this and other complementary methods are beyond the scope of this paper, they enable the empirical driving data discussed in this paper to provide utility for better understanding high-severity collision risk.

**Comparison benchmarks**

Human driver collision rates have been widely discussed as providing a benchmark for AVs [12,13]. However, ample care must be taken when choosing the benchmarks for comparison. The data in this paper includes all events involving actual or simulated contact between the AV and another object. By including low-speed events involving non-police-reportable contact (e.g. a less than 2 mph vehicle-to-vehicle contact or a pedestrian walking into the side of a stationary vehicle), the scope of events is considerably greater than the scope of police-reported or insurance-reported collisions commonly used to generate performance baselines. In a 2010 publication [14], NHTSA estimated 60% of property-damage-only accidents are not reported to the police. Furthermore, this estimate (60%) does not include no-damage contact events, as are included in this paper. As such, comparing the data presented in this paper to police-reported collision numbers is not an apt comparison. Obtaining reliable event counts that include such minor events typically requires analysis of naturalistic driving data.

Although Waymo has found the collision frequencies observed in this data to compare favorably to analogous frequencies observed in naturalistic driving studies [15], such comparisons are very challenging to perform validly. This is not only due to statistical variability but, more importantly, due to systematic uncertainties arising from the ODD-specificity of our data (e.g., road speed distribution and traffic density), inherent limitations of simulation, and assumptions in human response models. This cautious view of otherwise promising data is reflected in Waymo's overall approach to safety, which avoids a singular focus on collision frequencies derived from on-road data, and instead considers them as one methodology among several that ensure the safety of the Waymo Driver [1].

### 4.3 Limitations and Future Work

The data presented in this paper is from Waymo's ride hailing operations in the metro Phoenix area, and the scale and mix of driverless and self-driving with trained operators miles reported reflects Waymo's conservative approach to driverless deployments. Limitations related to the statistical power of the mileages reported have been discussed in the above section on aggregate collision frequencies. This section includes discussion of other limitations of the methodology and observations regarding the context of this paper and opportunities for future work.

**Limitations of counterfactual simulations**

The results in this paper share limitations common to all counterfactual simulations. Due to the nature of human agent behavior, disengagement simulations are not definitive: counterfactual simulations predict what could have occurred, but cannot definitively predict exactly what would have occurred. This is particularly relevant for collisions, which are rare events that often are the result of off-nominal behavior from one or more roadway users. As a result, had the driver not disengaged, some of the reported simulated collisions may not have actually occurred (e.g. other agents may have behaved differently). Conversely, other events that, in simulation did not result in contact, may have actually resulted in collisions (e.g. if the other agent had been distracted at the critical moment). As previously noted, counterfactual methods are evolving. Although more complex, a potential improvement could be to assign probabilistic counterfactual outcomes to account for a range of possible human behaviors and responses. Waymo therefore takes a cautious approach to interpreting both the outcomes of individual collisions and aggregate performance metrics, and considers them in the context of other indicators of AV performance [1].

**Secondary collision in simulated events**

The severities ascribed to the simulated collisions are based on the single impact depicted in the simulation. Owing to complexities in accurately modeling post-impact vehicle dynamics (which may or may not involve subsequent steering and braking maneuvers from the other vehicle), the outcome of any secondary collisions that might occur subsequent to the simulated event are not explicitly modeled. In Waymo's ODD, the vast majority of primary vehicle-to-vehicle collisions (99% for all collisions, 95% for fatal collisions) included in police-reported crash databases involve either a single vehicle-to-vehicle collision event or a subsequent collision event of equal or lesser severity. The method used in this paper therefore captures the vast majority but not all of the severity risk from each collision.

**Interpreting disengage performance**

Care should be taken in drawing conclusions based on the collision-avoidance performance of Waymo's trained operators during disengagements, which for the reasons described below, is not predictive of the collision-avoidance performance of the overall population of human drivers. The primary function of Waymo's trained operators is to ensure the safe operation of the AV through disengagements. Waymo vehicle operators are selected from a subset of the driving population with good driving records and receive instruction specific to Waymo AVs, defensive driving training, and education regarding fatigue. When operating a vehicle, strict rules are in place regarding handheld devices including cell phones and operators are continually monitored for signs of drowsiness. Unlike drivers in human-driven vehicles, while the AV is in self-driving mode, Waymo's trained operators do not execute navigation, path planning, or control tasks, but instead are focused on monitoring the environment and the Waymo Driver's response to it. Trained vehicle operators are therefore able to focus their full attention on being ready to disengage and execute collision avoidance, and their performance at this task is expected to be superior to that of a human in a traditional driving role.

**Future work**

We expect and invite other safety researchers to review the events and mileages presented here and make their own findings regarding the safety performance of Waymo's operations demonstrated in this data. For example, we expect that data from naturalistic driving studies may offer additional perspective on the results presented in this paper. However, considerable data sampling efforts may be needed to ensure comparable data sets (e.g. due to ODD specificity).

## 5. Conclusion

This paper provides a detailed analysis of every actual and simulated, counterfactual ("what if") collision or contact that was collected from more than 6.1 million miles of fully automated driving in Waymo's metro Phoenix ODD. This includes operations with a trained operator behind the steering wheel from calendar year 2019 and 65,000 miles of driverless operation without a human behind the steering wheel from 2019 and the first nine months of 2020. Taken together, these 47 lower severity (S0 and S1) events (18 actual and 29 simulated, one during driverless operation) show significant contribution from other agents, namely human-related deviations from traffic rules and safe driving performance. Nearly all the actual and simulated events involved one or more road rule violations or other incautious behavior by another agent, including all eight of the most severe events involving actual or expected airbag deployment.

The frequency of challenging events that were induced by incautious behaviors of other drivers serves as a clear reminder of the challenges in collision avoidance so long as AVs share roadways with human drivers. Statistics regarding the high percentage of human collisions that are attributed to human error may lead to inflated expectations of the potential immediate safety benefits of AVs. AVs will share roads with human drivers for the foreseeable future, and significant numbers of collisions due to human driver errors that are simply unavoidable should be expected during this period.

Due to the typology of those collisions initiated by other actors as well as the Waymo Driver's proficiency in avoiding certain collision modes, the data presented shows a significant shift in the relative distributions of collision types as compared to national crash statistics for human drivers. For example, the Waymo Driver experienced zero actual or simulated collision-relevant contacts in the NHTSA "road departure, fixed object, rollover" single-vehicle collision typology (27% of all US roadway fatalities). In an additional example, while rear-end collisions are one of the most common collision modes for human drivers, the Waymo Driver only recorded a single front-to-rear striking collision (simulated) and that event involved an agent cutting in and immediately braking (consistent with antagonistic motive).

This is the first time that information on every actual and simulated collision or contact has been shared for millions of miles of automated driving. Data from more than 6.1 million miles of driving (representing over 500 years of driving for the average U.S. licensed driver) provides sufficient statistical signal to detect moderate-to-large differences in S0 and S1 event frequencies. However, as discussed in *Waymo's Safety Methodologies and Safety Readiness Determinations*[1] our assessment of AV safety uses multiple complementary methodologies, including simulation and closed course testing, which allows for comprehensive testing, including rare scenarios beyond those encountered in this dataset. As AV fleets and their mileage continue to grow, so will our understanding of their safety impact. This data only represents the performance of the Waymo Driver at a snapshot in time, and the performance of the Waymo Driver is continually improving. Therefore, the most significant long-term contributions of this paper will likely not be the actual data shared, but the example set by publicly sharing this type of safety performance data and the dialogs that this paper fosters.